\documentclass[pmlr]{jmlr}

\usepackage{longtable}
\usepackage{booktabs}
\usepackage{caption}
\usepackage{float}

\makeatletter
 \let\Ginclude@graphics\@org@Ginclude@graphics 
\jmlrvolume{106}
\jmlryear{2019}
\jmlrworkshop{Machine Learning for Healthcare}

\title[Multimodal Machine Learning for Automated ICD Coding]{Multimodal Machine Learning for Automated ICD Coding}

\author{\Name{Keyang Xu}$^{\spadesuit}$ \Email{xky0714@gmail.com}\\
\Name{Mike Lam}, \Name{Jingzhi Pang}, \Name{Xin Gao}, \Name{Charlotte Band}$^{\spadesuit}$ \\
\Name{Piyush Mathur}, \Name{Frank Papay}, \Name{Ashish K. Khanna}, \Name{Jacek B. Cywinski}, \Name{Kamal Maheshwari}$^{\triangle}$\\
\Name{Pengtao Xie}$^{\spadesuit}$ \Email{pengtao.xie@petuum.com}\\
\Name{Eric P. Xing}$^{\spadesuit}$ \Email{eric.xing@petuum.com}\\
\addr $^{\spadesuit}$Petuum Inc, Pittsburgh, PA, USA\\
$^{\triangle}$Cleveland Clinic, Cleveland, Ohio USA}

\begin{document}

\maketitle

\begin{abstract}
    
  This study presents a multimodal machine learning model to predict ICD-10 diagnostic codes. We developed separate machine learning models that can handle data from different modalities, including unstructured text, semi-structured text and structured tabular data. We further employed an ensemble method to integrate all modality-specific models to generate ICD codes. Key evidence was also extracted to make our prediction more convincing and explainable.

    We used the Medical Information Mart for Intensive Care III (MIMIC-III) dataset to validate our approach. For ICD code prediction, our best-performing model (micro-F1 = 0.7633, micro-AUC = 0.9541)  significantly outperforms other baseline models including TF-IDF (micro-F1 = 0.6721, micro-AUC = 0.7879) and Text-CNN model (micro-F1 = 0.6569, micro-AUC = 0.9235). For interpretability, our approach achieves a Jaccard Similarity Coefficient (JSC) of 0.1806 on text data and 0.3105 on tabular data, where well-trained physicians achieve 0.2780 and 0.5002 respectively.
    
\end{abstract}

\section{Introduction}
\paragraph{Clinical Relevance} The International Classification of Diseases (ICD), endorsed by the World Health Organization (WHO), is a medical classification list of codes for diagnoses and procedures\footnote{https://www.cdc.gov/nchs/icd/icd10cm.htm}. ICD codes have been adopted widely by physicians and other health care providers for reimbursement, storage, and retrieval of diagnostic information (\cite{nadathur2010maximising}, \cite{bottle2008intelligent}).
The process of assigning ICD codes to a patient visit is time-consuming and error-prone. Clinical coders need to extract key information from Electronic Medical Records (EMRs) and assign correct codes based on category, anatomic site, laterality, severity, and etiology (\cite{quan2005coding}). 
The amount of information and complex hierarchy greatly increase the difficulty. Coding errors can further cause billing mistakes, claim denials, and underpayment~(\cite{adams2002addressing}). Therefore, an automatic and robust ICD coding process is in great demand.

EMRs store data in different modalities, such as unstructured text, semi-structured text, and structured tabular data~(\cite{bates2003proposal}). Unstructured text data includes notes taken by physicians, nursing notes, lab reports, test reports, and discharge summaries. Semi-structured text data refers to a list of structured phrases and unstructured sentences that describe diagnoses written by physicians. Structured tabular data contains prescriptions and clinical measurements such as vital signs, lab test results, and microbiology test results (\cite{leediagnosis}, \cite{lipton2015learning}, \cite{parthiban2012applying}). How to leverage all information from large-volume EMR data is a non-trivial task. Besides, in clinical practice, providing predictions with black-box machine learning models is not convincing for physicians and insurance companies. How to provide evidence related to predicted ICD codes is also an important task. 

\paragraph{Technical Significance}
In this work, we developed an ensemble-based approach that integrates three modality-specific models to fully exploit complementary information from multiple data sources and boost the predictive accuracy of ICD codes. We also explored methods to produce interpretable and explainable predictions.  The primary contribution of this study is twofold:
\begin{itemize}
    \item  We applied NLP and multimodal machine learning to predict ICD diagnostic codes, achieving state-of-the-art accuracy. The complementary nature of multimodal data makes our model more robust and accurate. In addition, we effectively addressed data imbalance issues, which is a very general problem for ICD code prediction. 
    \item  We proposed approaches to interpret predictions for unstructured text and structured tables, which make our prediction more explainable and convincing to physicians.
\end{itemize}

\section{Related Work}

Many researchers have explored the topic of automatic ICD coding using text data in recent years (\cite{stanfill2010systematic}).~\cite{larkey1996combining} considered it as a single-label classification on patient discharge summaries with multiple classifiers. ~\cite{kavuluru2015empirical} treated it as a multi-label classification problem~(\cite{tsoumakas2007multi}), and developed ranking approaches after text feature engineering and selection using EMR data. ~\cite{koopman2015automatic} developed a classification system combining Support Vector Machine (SVM) and rule-based methods to identify four high-impact diseases with patient death certificates. It is also common to see unsupervised and semi-supervised strategies applied in predicting ICD codes.~\cite{scheurwegs2017assigning} proposed an unsupervised medical concept extraction approach using an unlabeled corpus to frame clinical text into a list of concepts, which helps with ICD code prediction. Deep learning models are also widely used~(\cite{shickel2018deep}).~\cite{duarte2018deep} developed a recurrent neural model and~\cite{mullenbach2018explainable} adopted a CNN-based model with a per-label attention mechanism  to assign ICD codes based on the free-text description.~\cite{shi2017towards} used a neural architecture with Long Short-Term Memory (LSTM) and an attention mechanism that takes diagnostic descriptions as input to predict ICD codes. Similarly,~\cite{xie2018neural} further explored tree-of-sequences LSTM encoding and adversarial learning to improve the prediction results. 

\section{Methods}
 In this section, we will introduce our approach for predicting ICD-10 codes for a patient visit, as well as for identifying the evidence to make our predictions explainable. 

\begin{figure}[t]
\centering
\includegraphics[width=0.95\columnwidth]{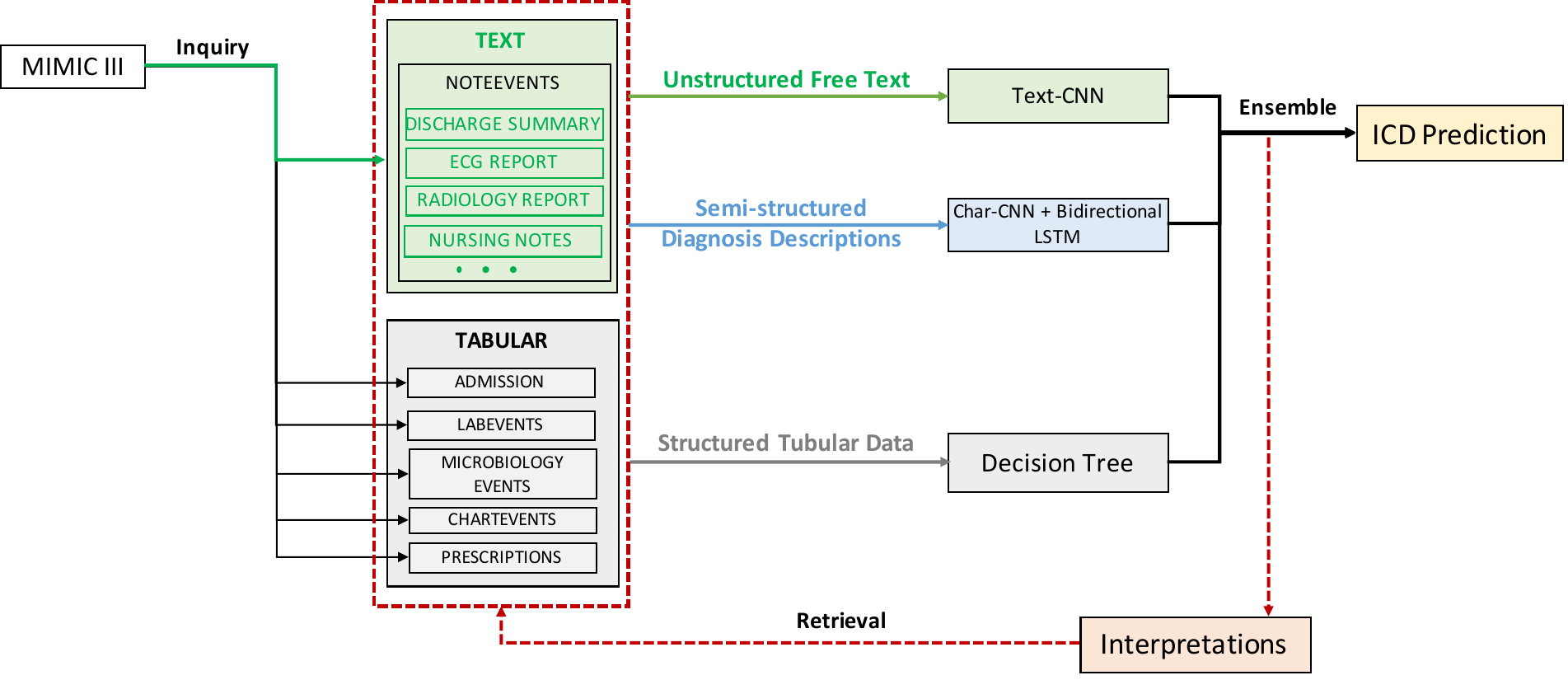}
\caption{Model architecture for ICD code prediction based on 
multimodal data, where each prediction is interpreted using retrieval-based methods.}
\label{fig:workflow}
\end{figure}

We developed an ensemble-based approach as shown in Figure~\ref{fig:workflow}, which integrates modality-specific models. For unstructured text, we applied the Text Convolutional Neural Network (Text-CNN) model~(\cite{DBLP:journals/corr/Kim14f}) for multi-label classification. For semi-structured text data, we constructed a deep learning model to analyze semantic similarities between diagnoses written by physicians and ICD code descriptions. For tabular data, our approach transformed numeric features into binary features and fed them to a decision tree~(\cite{Quinlan:1986}) to classify the ICD codes. During testing, our model ensembled the three aforementioned models on different modalities for improving prediction accuracy~(\cite{dietterich2000ensemble}). To make our predictions explainable, key evidence was retrieved from raw data and presented for examination.

\subsection{Data Structure Description}
The dataset used for this study is MIMIC-III~(\cite{johnson2016mimic}), which contains approximately 58,000 hospital admissions of 47,000 patients who stayed in the ICU of the Beth Israel Deaconess Medical Center between 2001 and 2012. The original diagnostic codes are ICD-9 codes. We mapped them to ICD-10 codes, as they are more widely adopted in today's clinical practices. ICD-9 codes with multiple corresponding ICD-10 codes are removed to avoid ambiguity. Based on clinical meaningfulness, 32 ICD codes that are top 50 frequent in both MIMIC-III and a national hospital in the U.S. were selected (see details in Table~\ref{tab:32 icd codes} in Appendix). Patient admissions that have ground truth labels of these 32 codes and at least one discharge summary were selected. In total we got 44,659 admissions, covering about 77\% of the whole dataset. Six major tables from the MIMIC-III dataset were selected: 
\begin{itemize}
  \item ADMISSIONS contains all information regarding a patient admission, including a preliminary diagnosis.
  \item LABEVENTS contains all laboratory measurements.
  \item PRESCRIPTIONS contains medications related to order entries.
  \item MICROBIOLOGYEVENTS contains microbiology information such as whether an organism tested negative or positive in the culture.
  \item CHARTEVENTS contains all charted data including patients' routine vital signs and other information related to their health.
  \item NOTEEVENTS contains all notes including nursing and physician notes, echocardiography reports, and discharge summaries. 
\end{itemize}

\subsection{Classification Based on Unstructured Text}

In this section, we will discuss how our model predicts diagnostic codes based on unstructured text from NOTEEVENTS. Our approach includes two steps: data pre-processing and deep learning-based classification.

The pre-processing step aims to provide a clean and standardized input for the deep learning model. It applied a standardized pipeline including tokenization and word normalization. Tokens with a frequency of less than ten were removed. At test time, out-of-vocabulary words were considered to be a special token ``UNK''.

The processed text was fed into a multi-label classification model for ICD code prediction. We denote the ICD code set as $\mathcal{Y} = \{1, 2, ..., C\}$. Given a text input $X$ which is a sequence of tokens, our goal is to select a subset of codes $Y \subseteq \mathcal{Y}$ that is most relevant to $X$. A Text-CNN model~(\cite{DBLP:journals/corr/Kim14f}) was applied to achieve this. This model represents tokens in $X$ using word embeddings~(\cite{mikolov2013efficient,mikolov2013distributed}), and applies a convolution layer on top of the embeddings to capture the temporal correlations among adjacent tokens. 
To learn the Text-CNN model, the objective is to minimize the average cross-entropy (CE) loss~(\cite{shore1980axiomatic}) for each ICD code class, defined as,
\begin{equation}
\label{eq:cross_entropy}
\mathcal{L}_{CE} =  - \frac{1}{N C} \Bigg[ \sum_{i=1}^{N} \sum_{j=1}^{C} I_{ij} \log P_{ij} + (1 - I_{ij}) \log \big(1 - P_{ij}\big)\Bigg],  
\end{equation}
where $N$ and $C$ are the number of training examples (patients' admissions) and unique ICD code. $I_{ij}$ is a binary label representing whether example $i$ is assigned with code $j$ and $P_{ij}$ is the probability (predicted by Text-CNN) indicating the likelihood that code $j$ is relevant to example $i$.

\begin{figure}[t]
\centering
\includegraphics[width=0.90\columnwidth]{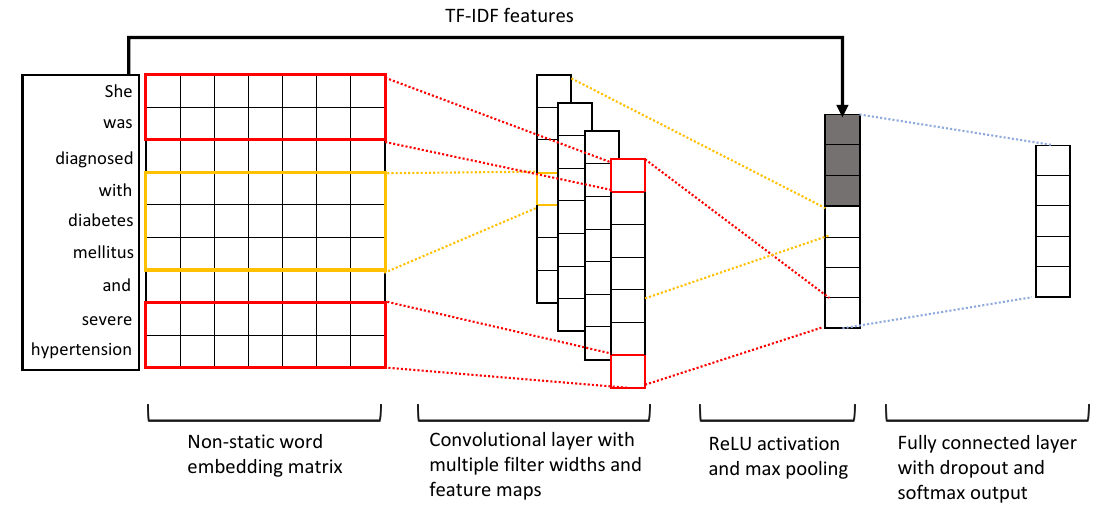}
\caption{A modified Text-CNN which uses clinical notes as inputs. The model that utilizes TF-IDF features is denoted as Text-TF-IDF-CNN.}
\label{fig:cnn}
\vspace{-1em}
\end{figure}

In practice, clinical guidelines developed by human experts are used to guide diagnoses. Medical knowledge from the guidelines can be leveraged by machine learning models to improve prediction accuracy. To achieve it, keywords and phrases were extracted from unstructured guidelines to generate TF-IDF feature vectors~(\cite{ramos2003using}). These features were fed into the fully connected layer of Text-CNN, as additional inputs. We refer to this modified model as Text-TF-IDF-CNN. The detailed network structure is shown in Figure \ref{fig:cnn}.

Training samples for 32 ICD code classes are not evenly distributed —-- some classes have many samples while others have few. Such data imbalance issue often significantly compromises the performance of machine learning models. To address it, we applied Label Smoothing Regularization (LSR)~(\cite{DBLP:journals/corr/SzegedyVISW15}). 
LSR prevents our classifier from being too certain about labels during training and thus avoids overfitting. The LSR approach is able to improve the generalization of our model. More specifically, as defined in \eqref{eq:cross_entropy},  $I_{ij} \in  \{0, 1\}$ and the average cross-entropy loss is minimized when $P_{ij}=I_{ij}$. LSR was implemented by replacing the truth label $I_{ij}$ with the linear combination between the ground truth label $I_{ij}$ and a sample drawn from a prior distribution $u(j)$, with weights $1-\epsilon$ and $\epsilon$. Here prior distribution is defined as $u(j)= {1}/{C}$. We sampled $\epsilon$ from a beta distribution in each iteration during training, where $\epsilon \sim \text{Beta}(\alpha, \alpha)$ and $\alpha \in (0, \infty)$ is the hyperparameter for Beta distribution. As a result, the smoothed label of class $j$ for sample $i$ is described as,
\begin{equation}\label{eq:ls}
\widetilde{I}_{ij} = (1-\epsilon) \cdot I_{ij} + \epsilon \cdot \frac{1}{C}
\end{equation}
The average cross-entropy (CE) loss is adjusted in Equation \eqref{eq:cross_entropy} by replacing ${I}_{ij}$ with the smoothed label $\widetilde{I}_{ij}$, to train our Text-CNN or Text-TF-IDF-CNN model.

\subsection{Ranking Based on Semi-structured Text}
Medical coders review the diagnosis descriptions taken by physicians in the form of textual phrases and sentences in the clinical notes or admissions table, then manually assign appropriate ICD codes by following the coding guidelines~(\cite{o2005measuring}) for billing purposes. For example, the description ``Acute on Chronic Kidney Failure'' is a strong signal for the ICD code N17.9 ``Acute Kidney Failure, unspecified'' because they share a close semantic similarity. 

We wanted to model this process and formulated this task as Diagnosis-based Ranking (DR) problem where latent representations of all the code descriptions are computed as points in a low-dimensional dense vector space. During inference, diagnosis descriptions are mapped to the same vector space and ICD codes are ranked based on their distance from the diagnosis vector. A neural network model that considers both the character- and word-level features was trained to represent the diagnoses and ICD code descriptions in the same space. To train the model, a triplet loss function is minimized~(\cite{cheng2016person}). It takes a triplet as input: \emph{diagnosis} example ($\mathbf{x}^{a}$), \emph{positive} code description example $\mathbf{x}^{p}$ and \emph{negative} code description example $\mathbf{x}^{n}$, respectively. The triplet loss defines a relative similarity between instances, by minimizing the distance between the positive pairs and maximizing the distance between the negative ones, defined as:
\begin{align}
    \label{eq:triplet}
     \mathcal{L}_{triplet} = \frac{1}{N} \Big[\sum_{i=1}^{N}\max\{d(\mathbf{x}^{a}_i, \mathbf{x}^{p}_i) - d (\mathbf{x}^{a}_i, \mathbf{x}^{n}_i) + \alpha, 0\}\Big],
\end{align}
where $d(\cdot)$ is the Euclidean distance 
$d(\mathbf{x}, \mathbf{y}) = \|\mathbf{x} - \mathbf{y}\|_{2}$ for $\mathbf{x}, \mathbf{y} \in \mathbb{R}^{d}$ and $\alpha$ is the margin. Here we took margin $\alpha$ as 1. The associated network structure is shown in Figure~\ref{fig:dr}. Pre-trained word embedding and Character-level Convolution Neural Network (Char-CNN) (\cite{zhang2015character}) are used to represent each word token. The Char-CNN layer creates representations such that tokens with a similar form like ``injection'' and ``injections'' are mapped closer to each other. Similarly, the word embeddings represent a semantic space, such that a token ``Hypertension'' and its abbreviation ``HTN'' map closer to each other. The word embedding matrix was pre-trained on PubMed\footnote{https://www.ncbi.nlm.nih.gov/pubmed/}  (\cite{mikolov2013distributed}), which contains abstracts of over 550,000 biomedical papers. Then, a bidirectional LSTM (\cite{huang2015bidirectional}) layer encodes the sequential context information followed by a max-pooling layer to compute final representations.

The MIMIC-III dataset does not have a one-to-one mapping of ICD code and diagnosis. To construct the triplets for training, we crawled ICD-10 codes and synonyms for their corresponding code description from online resources\footnote{https://www.icd10data.com/}. Then, we mapped all the synonyms for each ICD code as positive examples. To create high-quality negative examples, instead of randomly sampling, we computed the edit distance to extract n-grams that are similar to the code description. During inference, min-max normalization transformed the distance value to a range [0, 1]. If one admission has multiple diagnoses, the mean of features was computed to combine those normalized results.


\subsection{Classification based on Tabular Data}
Tabular data mainly includes four tables, including LABEVENTS, PRESCRIPTIONS, MICROBIOLOGYEVENTS, and CHARTEVENTS. 
\begin{itemize}
  \item LABEVENTS (LAB): It uses a binary value to indicate whether each laboratory value is abnormal. Furthermore, if a patient has multiple records of the same lab test, a majority vote strategy was adopted. Missing values were considered as normal. In our experiments, a total of 753 lab tests were used to construct features for classification.
  \item  CHARTEVENTS: Based on physicians' suggestion, we mainly focused on three measurements: Body Mass Index (BMI), heart rate, and blood pressure. A binary vector was used to denote whether a specific measurement result is within the normal range. We combined features from this table with LABEVENTS. 
  \item  PRESCRIPTIONS (MED): A binary vector was used to denote whether a medication is prescribed. Missing medications were considered as not being prescribed. Medications with a frequency of less than 50 were removed to reduce noise. A total of 1135 medications were used to construct binary features.
  \item  MICROBIOLOGYEVENTS (BIO): A binary feature was used to represent microbiology even --- whether an organism test is positive or negative. A total of 363 organisms were used to form feature vectors.
  
\end{itemize}

Our solution applied a decision tree~(\cite{Quinlan:1986}) as the classifier using binary features and leveraged a one-versus-all strategy ~(\cite{Rifkin:2004}) for multi-label classification. Higher weights were given to samples from minority classes to handle data imbalance.

\subsection{Model Ensemble}
During testing, our solution takes an ensemble of the trained models for predicting ICD codes. Specifically, for each class $j$, the final predicted probability $P_j$ is a weighted sum of probabilities predicted by individual models:
\begin{align}
    P_j = \sum_{k=1}^{K}{\alpha_k P_{j}^{(k)}},
\end{align}
where $P_{j}^{(k)}$ is the probability predicted by model $k$ on class $j$, and $K$ is the total number of models. The weight parameters $\{\alpha_k\}_{k=1}^K$ were tuned by performing a grid search on the validation set. However, not every single patient visit is guaranteed to have all the data resources available.  
Hence in order to ensure all weights sum up to one, i.e., $\sum_{k=1}^{K}\alpha_k = 1$, if the $k$-th  predictor is missing, its weight $\alpha_k$ will be given to the Text-CNN or Text-TF-IDF-CNN model.

\subsection{Methods for Interpretation}
Two separate methods were adopted to extract the important textual and tabular features that have the highest influence on the final ICD code predictions. 

\subsubsection{Text Interpretability}
After our model predicts a specific ICD code $y \in \{1, 2, ..., C\}$, it is desirable to identify key phrases that lead to such a prediction from the textual input. To capture the association between a word $w$ and an ICD code $y$, we extracted all the paths connecting $w$ and $y$ from the trained neural network. For each path, we computed the influence score by multiplying the values of all hidden units and the weights associated with all edges along this path. The scores of all paths were added up to measure the association strength between $w$ and $y$.
Consecutive words with non-zero scores were combined into phrases and then ranked by the maximum score. Top-ranked phrases were considered important signals for models to determine a specific ICD code. 

\subsubsection{Table Interpretability}
The nature of structured tabular features makes the method inherently different from those using text data. 
Therefore, the Local Interpretable Model-Agnostic Explanation (LIME) method~(\cite{DBLP:journals/corr/RibeiroSG16}) was adopted to tackle this problem.
LIME computes a score for each feature of an instance that represents the importance of this feature in contributing to the model's final prediction. Instead of going through the trained model, LIME learns the weights of features by approximating the decision boundary locally (i.e. instance-specific) based on a simple model.

\section{Experiments}
In this section, we will introduce experiments for validating our approach and evaluation results for both classification and interpretability.

\subsection{Experimental Settings}
The data was randomly split into training, validation, and test sets containing 31,155, 4,484, and 9,020 admissions respectively. Admissions of the same patient were categorized into the same set. This prevents the model from memorizing information of a patient from the training set and leveraging that information to inflate performance on the test set. 

\subsection{Hyperparameters} 
Hyperparameters were tuned on the validation set and then applied to the test set, including: (1) For the Text-CNN and Text-TF-IDF-CNN model, kernel sizes of 2, 3, 4 were adopted and each kernel had 128 feature maps. Word embedding dimension was 256. Dropout rate was 0.1. L2-regularization coefficient was $10^{-4}$. Adam optimizer~(\cite{kingma2014adam}) with learning rate $10^{-3}$ was used. The batch size was 32; $\alpha$ was 0.3 for beta distribution in label smoothing regularization. (2) For diagnosis ranking, the character embedding size was 50. One layer of bi-directional LSTM with 100 hidden units was used for encoding. 

\begin{table}[t]
\centering 
\caption{Evaluations of classification for our models and baselines. LS stands for label smoothing. DR is the diagnosis-based ranking model. LAB contains both LABEVENTS and CHARTEVENTS features. Tabular Data (TD) includes Bio, Med, and Lab.}
\label{Model Performance}
\begin{tabular}{lllll}
\toprule
\multicolumn{1}{c}{{Methods}}                             & \multicolumn{2}{c}{F1} & \multicolumn{2}{c}{AUC} \\ \cline{2-5} 
\multicolumn{1}{c}{}  & \multicolumn{1}{c}{Macro} & \multicolumn{1}{c}{Micro} & \multicolumn{1}{c}{Macro} & \multicolumn{1}{c}{Micro}     \\ \hline
Word-TF-IDF    & .5277  & .6648   & .7249 & .7834   \\
\begin{tabular}[c]{@{}l@{}}Word-TF-IDF + MetaMap-TF-IDF\end{tabular}   & .5336  & .6721  & .7298   & .7879   \\
Keyword-TF-IDF    & .5678  & .7225   & .7375 & .7867   \\ \hline
DenseNet     & .5327     & .6621    & .8885 & .9228    \\ 
Text-CNN  & .5429  & .6569  & .8959 & .9235  \\ \hline
Text-CNN + LS    & .6054  & .6908  & .9010  & .9293 \\
Text-CNN + LS + DR   & .6097  & .6914  & .9029  & .9307    \\ 
Text-CNN + LS + Bio & .6108  & .6922 & .9076   & .9323    \\ 
Text-CNN + LS + Lab  & .6107    & .6925  & .9136   & .9366   \\
Text-CNN + LS + Med  & .6122   & .6928    & .9124  & .9379   \\ 
\begin{tabular}[c]{@{}l@{}}Text-CNN + LS + TD\end{tabular} & .6137   & .6929  & .9177   & .9406  \\
Text-CNN + LS + DR + TD & .6133   & .6921    & .9188    & .9416  \\ \hline
Text-TF-IDF-CNN + LS  & .6813  & .7632  & .9157 & .9420  \\
\begin{tabular}[c]{@{}l@{}} Text-TF-IDF-CNN + LS + DR + TD \end{tabular} & $\mathbf{.6867}$  & $\mathbf{.7633}$  & $\mathbf{.9337}$ & $\mathbf{.9541}$ \\ \bottomrule
\end{tabular}
\end{table}

\subsection{Evaluation Metrics}
\begin{itemize}
    \item \textbf{For Classification}, we used F1 and the Area Under the ROC Curve (AUC)~(\cite{goutte2005probabilistic, huang2005using}) as evaluation metrics. F1 score is the harmonic mean of precision and recall, and AUC score summarizes performances under different thresholds.  
    To better compute the average across different classes, we adopted micro-averages and macro-averages. Classes with more samples have larger weights for micro-averaged metrics but are treated equally for macro-averaged metrics. 
    \item \textbf{For Interpretability}, we used the Jaccard Similarity Coefficient (JSC)~(\cite{niwattanakul2013using}) to measure the overlap between two sets, which are our extracted evidence and physicians' annotations. It is defined as,
    \begin{align*}
        J(A,B) = \frac{|A \cap B|}{ |A \cup B|}
    \end{align*}
\end{itemize}

\subsection{Evaluation of Code Classification}
The overall performance for multi-label classification is shown in Table~\ref{Model Performance}. 
Word-TF-IDF and MetaMap-TF-IDF are baseline methods, where MetaMap extracts medical entities and links them to UMLS concepts~(\cite{aronson2001effective}). Keyword-TF-IDF only adopts TF-IDF values of keywords from clinical guidelines. We used gradient boosting decision tree~(\cite{Stanford2002GreedyFA}) model as the classifier. 

Most models in Table~\ref{Model Performance} were developed based on the CNN architecture. Vanilla Text-CNN model and DenseNet~(\cite{huang2017densely}) perform similarly with TF-IDF models on F1 but achieve better results on AUC.

Label Smoothing (LS) significantly improves the performance as it alleviates data imbalance issue, especially for F1 scores. TextCNN+DR also enhances Text-CNN performance on all four metrics. 

For Tabular Data (TD) that comprises BIO, LAB and MED, we observe improvements on all four metrics after ensembling any type with Text-CNN+LS. Incorporating TD, Text-CNN+LS+TD gets the best macro- and micro-F1 (0.6137 and 0.6929 respectively) among all models with Text-CNN structure. By further ensembling DR, our solution gets a higher AUC score but slightly lower F1 scores.
Text-CNN+LS+DR+TD improves 13\% on macro-F1 and 5.4\% on micro-F1 over Text-CNN, and 25.9\% on macro-F1 and 19.5\% on micro-F1 over Word-TF-IDF, which demonstrates the effectiveness of our proposed ensemble approach and the benefits of using multimodal data. 
 
\begin{figure}[t]
\centering
\includegraphics[width=0.95\columnwidth]{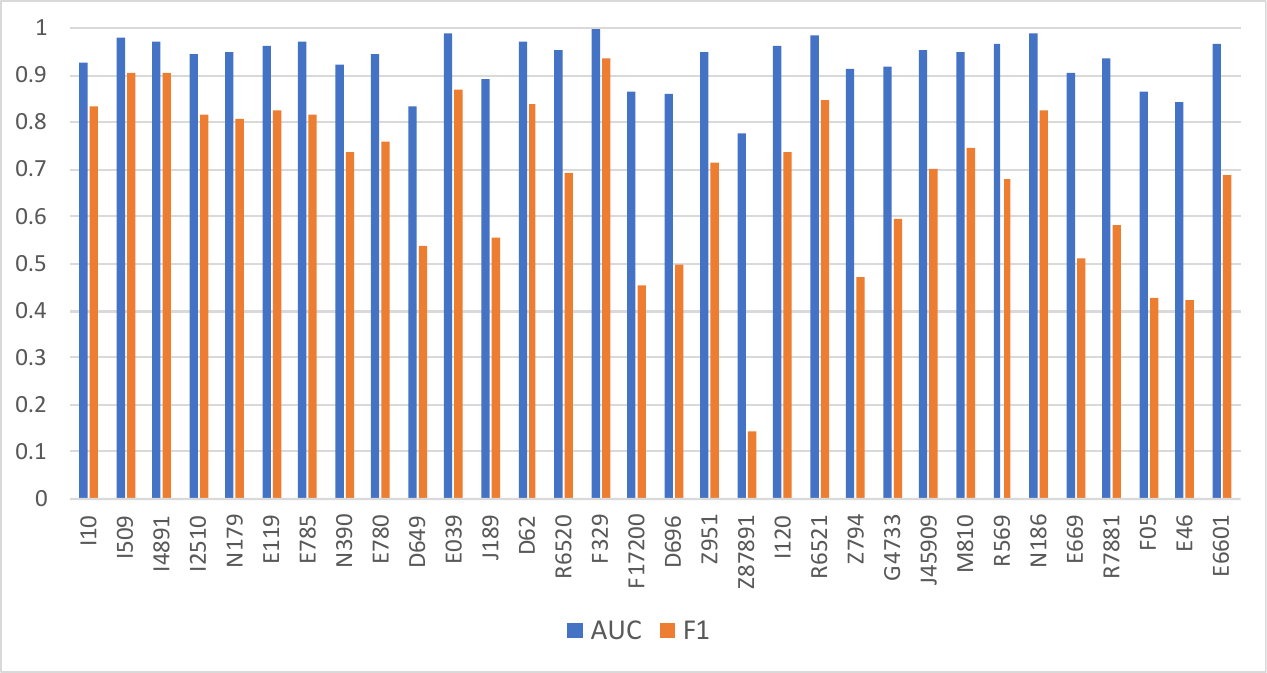}
\caption{Detailed AUC and F1 scores for 32 ICD codes for Text-TF-IDF-CNN + LS + TD + DR model. ICD codes are listed on the X axis by the training sample size, in descending order.} 
\label{fig:auc-f1}
\end{figure}

In addition, evaluation results show that the use of the most relevant subsets from clinical guidelines denoted as Text-TF-IDF-CNN can significantly increase performance. The combined model of Text-TF-IDF-CNN, LS, DR, and TD achieves 0.6867 Macro-F1, 0.7633 Micro-F1, 0.9337 Macro-AUC, and 0.9541 Micro-AUC, the highest scores among all models tested. Detailed evaluation results for 32 ICD classes are listed in Figure \ref{fig:auc-f1}.

\subsection{Interpretability Evaluation}
We collected a test set of 25 samples from five selected ICD-10 codes, including I10, I50.9, N17.9, E11.9, and D64.9. Annotations were obtained independently from three practicing physicians, who were asked to annotate key evidence from all data sources that determines the assignment of ICD-10 codes. The physicians have 15, 22 and 32 years of experience, respectively.

\subsubsection{Text-oriented evaluation}
We compare top-$k$ phrases extracted by our model with physicians' annotations. $k$ is the number of snippets the physician annotated. If tokenized phrases and annotations coincide above a certain threshold, they are considered overlapped.  This also applies when comparing overlap between physician's annotations. We find different physicians have different annotation styles. For instance, physicians 2 and 3 tend to highlight keywords while physician 1 tends to highlight sentences, resulting in fewer tokens obtained from physicians 2 and 3. Therefore, the average overlap score is as low as 0.2784 among highly-trained professionals, indicating that finding evidence from clinical notes to determine ICD codes is a non-trivial task.

\begin{table}[t]
\begin{center}
\caption{Jaccard Similarity Coefficient (JSC) between physicians and our model}
\label{Jaccard Similarity}
\begin{tabular}{cccccc}
\toprule
\begin{tabular}[c]{@{}l@{}}\end{tabular} & \begin{tabular}[c]{@{}l@{}}JSC\end{tabular} & \begin{tabular}[c]{@{}c@{}}Physician \\ 1 \end{tabular} & \begin{tabular}[c]{@{}c@{}}Physician \\ 2\end{tabular} & \begin{tabular}[c]{@{}c@{}}Physician \\ 3\end{tabular} & \begin{tabular}[c]{@{}c@{}}Our\\ Model\end{tabular} \\ \hline
  & \begin{tabular}[c]{@{}l@{}}Physician 1\end{tabular}   & -    & .3105  &  .1798 & .1477 \\ 
Text Data &\begin{tabular}[c]{@{}l@{}}Physician 2\end{tabular}   & .3105   & -  & .3450 & .2137  \\
&\begin{tabular}[c]{@{}l@{}}Physician 3\end{tabular}   & .1798   & .3450  & -  & .1905  \\ 
\hline
&\begin{tabular}[c]{@{}l@{}}Physician 1\end{tabular}   & -   & .5455  &  .4225 & .3867 \\ 
Tabular Data &\begin{tabular}[c]{@{}l@{}}Physician 2\end{tabular}   & .5455   & -  & .5325 & .2885  \\ 
&\begin{tabular}[c]{@{}l@{}}Physician 3\end{tabular}   & .4225   & .5325  & -  & .2564  \\ 
\bottomrule
\end{tabular}
\end{center}
\vspace{-1em}
\end{table}

Table~\ref{Jaccard Similarity} shows the overlap score between physicians' annotations and the outputs of our model. On average, our model obtains JSC of 0.1806. Our approach is designed to find n-grams carrying important signals. It can capture numerous phrases that are either directly related to a specific disease or that can provide insights for the final code prediction. 

\subsubsection{Table-oriented evaluation}

For each table, the $k$ most important features found by LIME were selected as evidence for each sample, where $k$ is the number of important features a physician annotated. We removed duplicates and computed the JSC. All results are shown in Table~\ref{Jaccard Similarity}.  We observe that the average agreement among physicians on tabular data is 0.5002, higher than that on text data. The average JSC between our model and physicians' annotations is  0.3105. In general, our model captures more features than physicians. Some of those extra features are informative for diagnoses even though they are not annotated by physicians. This is similar to the finding from text interpretability.

\section{Discussion}
\subsection{Code Classification}
The use of TF-IDF features makes the vector space model perform well on F1 score, indicating the importance of retrieving disease-related keywords in these classification tasks. 
Inspired by Keyword-TF-IDF which filters out noise and irrelevant features, we proposed Text-TF-IDF-CNN which incorporates cognitive TF-IDF features extracted from clinical guidelines to the fully-connected layer in our deep learning model. It shows the possibility of leveraging human domain knowledge to improve ML model performance in clinical settings. 

Performance improvements on macro-averaged metrics are greater than on micro-averaged metrics. One reason is that multimodal data mitigates the issue of limited inference from text data caused by a small training sample size. Additionally, given a physical examination and lab tests are mandatory for most inpatients, tabular data considerably benefits classes with small sample size. As we expand the prediction scope to cover a larger code list in the future, resolving data imbalance will play a bigger role in improving predictive accuracy.  

\subsection{Interpretability}
\label{interp_discussion}
For text data, key evidence was extracted based on the Text-TF-IDF-CNN model. For tabular data, key evidence was extracted based on LIME from LABEVENTS and PRESCRIPTIONS tables. 

For text data, among all ranked evidence extracted, our model is able to identify keywords highly relevant to a specific code. For example, ``DM II'' and ``Metformin'' were extracted for E11.9. Results reveal that our model is able to identify text snippets that are related to a specific disease, even though the relationship can be indirect. Take N17.9 for example, the snippet ``IVDU, states he used heroin, barbiturates, cocaine'' indicates drug abuse could be the cause of acute kidney failure.

For tabular data, valuable items can be extracted as well, especially where text evidence is not sufficient. For example, ``Glucose'' from LABEVENTS and ``Insulin'' from PRESCRIPTIONS were extracted as top evidence for E11.9, which were not found in top-ranked keywords from unstructured text.  

Interpretation results are affected by the performance of classification. For example, items extracted in classes with an AUC score above 0.9 are more relevant than those in class D64.9 with an AUC score around 0.7 (see Figure \ref{fig:auc-f1}).

Table~\ref{interpretability demonstration} in the Appendix shows some examples of evidence extracted from multiple modalities.

\subsection{Future Work}
We plan to improve our models in the following areas:
\begin{itemize}
    \item Enlarge the code list. Currently, we focus on 32 ICD-10 codes that are representative in hospitals and we will cover more minority classes.
    \item Reduce feature dimensions. Currently, feature dimensions for tabular data are very large, and some features are duplicated. Feature dimension reduction and duplicate removal may improve performance for both classification and interpretability. 
    \item Add human knowledge. Incorporating domain expertise proves helpful in improving the prediction. Exploration of new methods to better utilize existing knowledge besides using keyword matching can be another interesting research direction. 
\end{itemize}

\section{Conclusion}
We proposed a novel multimodal machine learning approach to predict ICD-10 diagnostic codes using EMRs. We developed separate models that handle data from different modalities, including unstructured text, semi-structured text, and structured tabular data. Experiments show that our ensembled model outperforms all the baseline methods in classification. Incorporating human knowledge into machine learning models to further enhance performance is proved effective, and we would like to explore more about it in our future work. In addition,
our interpretability method makes our prediction more explainable, transparent, and trustworthy. The capability to extract valuable information that includes but is not limited to code-specific text phrases, important lab test records and prescriptions, demonstrates great potential for use in real clinical practice.

\bibliography{reference}

\begin{thebibliography}{41}
\providecommand{\natexlab}[1]{#1}
\providecommand{\url}[1]{\texttt{#1}}
\expandafter\ifx\csname urlstyle\endcsname\relax
  \providecommand{\doi}[1]{doi: #1}\else
  \providecommand{\doi}{doi: \begingroup \urlstyle{rm}\Url}\fi

\bibitem[lar(1996)]{larkey1996combining}
Combining classifiers in text categorization.
\newblock In \emph{Proceedings of the 19th annual international ACM SIGIR
  conference on Research and development in information retrieval}, pages
  289--297. ACM, 1996.

\bibitem[Adams et~al.(2002)Adams, Norman, and Burroughs]{adams2002addressing}
Diane~L Adams, Helen Norman, and Valentine~J Burroughs.
\newblock Addressing medical coding and billing part ii: a strategy for
  achieving compliance. a risk management approach for reducing coding and
  billing errors.
\newblock \emph{Journal of the National Medical Association}, 94\penalty0
  (6):\penalty0 430, 2002.

\bibitem[Aronson(2001)]{aronson2001effective}
Alan~R Aronson.
\newblock Effective mapping of biomedical text to the umls metathesaurus: the
  metamap program.
\newblock In \emph{Proceedings of the AMIA Symposium}, page~17. American
  Medical Informatics Association, 2001.

\bibitem[Bates et~al.(2003)Bates, Ebell, Gotlieb, Zapp, and
  Mullins]{bates2003proposal}
David~W Bates, Mark Ebell, Edward Gotlieb, John Zapp, and HC~Mullins.
\newblock A proposal for electronic medical records in us primary care.
\newblock \emph{Journal of the American Medical Informatics Association},
  10\penalty0 (1):\penalty0 1--10, 2003.

\bibitem[Bottle and Aylin(2008)]{bottle2008intelligent}
Alex Bottle and Paul Aylin.
\newblock Intelligent information: a national system for monitoring clinical
  performance.
\newblock \emph{Health services research}, 43\penalty0 (1p1):\penalty0 10--31,
  2008.

\bibitem[Cheng et~al.(2016)Cheng, Gong, Zhou, Wang, and Zheng]{cheng2016person}
De~Cheng, Yihong Gong, Sanping Zhou, Jinjun Wang, and Nanning Zheng.
\newblock Person re-identification by multi-channel parts-based cnn with
  improved triplet loss function.
\newblock In \emph{Proceedings of the IEEE Conference on Computer Vision and
  Pattern Recognition}, pages 1335--1344, 2016.

\bibitem[Dietterich(2000)]{dietterich2000ensemble}
Thomas~G Dietterich.
\newblock Ensemble methods in machine learning.
\newblock In \emph{International workshop on multiple classifier systems},
  pages 1--15. Springer, 2000.

\bibitem[Duarte et~al.(2018)Duarte, Martins, Pinto, and Silva]{duarte2018deep}
Francisco Duarte, Bruno Martins, C{\'a}tia~Sousa Pinto, and M{\'a}rio~J Silva.
\newblock Deep neural models for icd-10 coding of death certificates and
  autopsy reports in free-text.
\newblock \emph{Journal of biomedical informatics}, 80:\penalty0 64--77, 2018.

\bibitem[Goutte and Gaussier(2005)]{goutte2005probabilistic}
Cyril Goutte and Eric Gaussier.
\newblock A probabilistic interpretation of precision, recall and f-score, with
  implication for evaluation.
\newblock In \emph{European Conference on Information Retrieval}, pages
  345--359. Springer, 2005.

\bibitem[Huang et~al.(2017)Huang, Liu, van~der Maaten, and
  Weinberger]{huang2017densely}
Gao Huang, Zhuang Liu, Laurens van~der Maaten, and Kilian~Q. Weinberger.
\newblock Densely connected convolutional networks.
\newblock In \emph{2017 {IEEE} Conference on Computer Vision and Pattern
  Recognition, {CVPR} 2017, Honolulu, HI, USA, July 21-26, 2017}, pages
  2261--2269, 2017.

\bibitem[Huang and Ling(2005)]{huang2005using}
Jin Huang and Charles~X Ling.
\newblock Using auc and accuracy in evaluating learning algorithms.
\newblock \emph{IEEE Transactions on knowledge and Data Engineering},
  17\penalty0 (3):\penalty0 299--310, 2005.

\bibitem[Huang et~al.(2015)Huang, Xu, and Yu]{huang2015bidirectional}
Zhiheng Huang, Wei Xu, and Kai Yu.
\newblock Bidirectional lstm-crf models for sequence tagging.
\newblock \emph{arXiv preprint arXiv:1508.01991}, 2015.

\bibitem[Johnson et~al.(2016)Johnson, Pollard, Shen, Li-wei, Feng, Ghassemi,
  Moody, Szolovits, Celi, and Mark]{johnson2016mimic}
Alistair~EW Johnson, Tom~J Pollard, Lu~Shen, H~Lehman Li-wei, Mengling Feng,
  Mohammad Ghassemi, Benjamin Moody, Peter Szolovits, Leo~Anthony Celi, and
  Roger~G Mark.
\newblock Mimic-iii, a freely accessible critical care database.
\newblock \emph{Scientific data}, 3:\penalty0 160035, 2016.

\bibitem[Kavuluru et~al.(2015)Kavuluru, Rios, and Lu]{kavuluru2015empirical}
Ramakanth Kavuluru, Anthony Rios, and Yuan Lu.
\newblock An empirical evaluation of supervised learning approaches in
  assigning diagnosis codes to electronic medical records.
\newblock \emph{Artificial intelligence in medicine}, 65\penalty0 (2):\penalty0
  155--166, 2015.

\bibitem[Kim(2014)]{DBLP:journals/corr/Kim14f}
Yoon Kim.
\newblock Convolutional neural networks for sentence classification.
\newblock \emph{CoRR}, abs/1408.5882, 2014.

\bibitem[Kingma and Ba(2014)]{kingma2014adam}
Diederik~P Kingma and Jimmy Ba.
\newblock Adam: A method for stochastic optimization.
\newblock \emph{arXiv preprint arXiv:1412.6980}, 2014.

\bibitem[Koopman et~al.(2015)Koopman, Zuccon, Nguyen, Bergheim, and
  Grayson]{koopman2015automatic}
Bevan Koopman, Guido Zuccon, Anthony Nguyen, Anton Bergheim, and Narelle
  Grayson.
\newblock Automatic icd-10 classification of cancers from free-text death
  certificates.
\newblock \emph{International journal of medical informatics}, 84\penalty0
  (11):\penalty0 956--965, 2015.

\bibitem[Lee and Muis()]{leediagnosis}
Jeong~Min Lee and Aldrian~Obaja Muis.
\newblock Diagnosis code prediction from electronic health records as
  multilabel text classification: A survey.

\bibitem[Lipton et~al.(2015)Lipton, Kale, Elkan, and
  Wetzel]{lipton2015learning}
Zachary~C Lipton, David~C Kale, Charles Elkan, and Randall Wetzel.
\newblock Learning to diagnose with lstm recurrent neural networks.
\newblock \emph{arXiv preprint arXiv:1511.03677}, 2015.

\bibitem[Mikolov et~al.(2013{\natexlab{a}})Mikolov, Chen, Corrado, and
  Dean]{mikolov2013efficient}
Tomas Mikolov, Kai Chen, Greg Corrado, and Jeffrey Dean.
\newblock Efficient estimation of word representations in vector space.
\newblock \emph{arXiv preprint arXiv:1301.3781}, 2013{\natexlab{a}}.

\bibitem[Mikolov et~al.(2013{\natexlab{b}})Mikolov, Sutskever, Chen, Corrado,
  and Dean]{mikolov2013distributed}
Tomas Mikolov, Ilya Sutskever, Kai Chen, Greg~S Corrado, and Jeff Dean.
\newblock Distributed representations of words and phrases and their
  compositionality.
\newblock In \emph{Advances in neural information processing systems}, pages
  3111--3119, 2013{\natexlab{b}}.

\bibitem[Mullenbach et~al.(2018)Mullenbach, Wiegreffe, Duke, Sun, and
  Eisenstein]{mullenbach2018explainable}
James Mullenbach, Sarah Wiegreffe, Jon Duke, Jimeng Sun, and Jacob Eisenstein.
\newblock Explainable prediction of medical codes from clinical text.
\newblock \emph{arXiv preprint arXiv:1802.05695}, 2018.

\bibitem[Nadathur(2010)]{nadathur2010maximising}
Shyamala~G Nadathur.
\newblock Maximising the value of hospital administrative datasets.
\newblock \emph{Australian Health Review}, 34\penalty0 (2):\penalty0 216--223,
  2010.

\bibitem[Niwattanakul et~al.(2013)Niwattanakul, Singthongchai, Naenudorn, and
  Wanapu]{niwattanakul2013using}
Suphakit Niwattanakul, Jatsada Singthongchai, Ekkachai Naenudorn, and
  Supachanun Wanapu.
\newblock Using of jaccard coefficient for keywords similarity.
\newblock In \emph{Proceedings of the International MultiConference of
  Engineers and Computer Scientists}, volume~1, 2013.

\bibitem[O'malley et~al.(2005)O'malley, Cook, Price, Wildes, Hurdle, and
  Ashton]{o2005measuring}
Kimberly~J O'malley, Karon~F Cook, Matt~D Price, Kimberly~Raiford Wildes,
  John~F Hurdle, and Carol~M Ashton.
\newblock Measuring diagnoses: Icd code accuracy.
\newblock \emph{Health services research}, 40\penalty0 (5p2):\penalty0
  1620--1639, 2005.

\bibitem[Parthiban and Srivatsa(2012)]{parthiban2012applying}
G~Parthiban and SK~Srivatsa.
\newblock Applying machine learning methods in diagnosing heart disease for
  diabetic patients.
\newblock \emph{International Journal of Applied Information Systems (IJAIS)},
  3:\penalty0 2249--0868, 2012.

\bibitem[Quan et~al.(2005)Quan, Sundararajan, Halfon, Fong, Burnand, Luthi,
  Saunders, Beck, Feasby, and Ghali]{quan2005coding}
Hude Quan, Vijaya Sundararajan, Patricia Halfon, Andrew Fong, Bernard Burnand,
  Jean-Christophe Luthi, L~Duncan Saunders, Cynthia~A Beck, Thomas~E Feasby,
  and William~A Ghali.
\newblock Coding algorithms for defining comorbidities in icd-9-cm and icd-10
  administrative data.
\newblock \emph{Medical care}, pages 1130--1139, 2005.

\bibitem[Quinlan(1986)]{Quinlan:1986}
J.~R. Quinlan.
\newblock Induction of decision trees.
\newblock \emph{Mach. Learn.}, 1\penalty0 (1):\penalty0 81--106, March 1986.
\newblock ISSN 0885-6125.
\newblock \doi{10.1023/A:1022643204877}.

\bibitem[Ramos et~al.(2003)]{ramos2003using}
Juan Ramos et~al.
\newblock Using tf-idf to determine word relevance in document queries.
\newblock In \emph{Proceedings of the first instructional conference on machine
  learning}, volume 242, pages 133--142, 2003.

\bibitem[Ribeiro et~al.(2016)Ribeiro, Singh, and
  Guestrin]{DBLP:journals/corr/RibeiroSG16}
Marco~T{\'{u}}lio Ribeiro, Sameer Singh, and Carlos Guestrin.
\newblock "why should {I} trust you?": Explaining the predictions of any
  classifier.
\newblock \emph{CoRR}, abs/1602.04938, 2016.

\bibitem[Rifkin and Klautau(2004)]{Rifkin:2004}
Ryan Rifkin and Aldebaro Klautau.
\newblock In defense of one-vs-all classification.
\newblock \emph{J. Mach. Learn. Res.}, 5:\penalty0 101--141, December 2004.
\newblock ISSN 1532-4435.

\bibitem[Scheurwegs et~al.(2017)Scheurwegs, Luyckx, Luyten, Goethals, and
  Daelemans]{scheurwegs2017assigning}
Elyne Scheurwegs, Kim Luyckx, L{\'e}on Luyten, Bart Goethals, and Walter
  Daelemans.
\newblock Assigning clinical codes with data-driven concept representation on
  dutch clinical free text.
\newblock \emph{Journal of biomedical informatics}, 69:\penalty0 118--127,
  2017.

\bibitem[Shi et~al.(2017)Shi, Xie, Hu, Zhang, and Xing]{shi2017towards}
Haoran Shi, Pengtao Xie, Zhiting Hu, Ming Zhang, and Eric~P Xing.
\newblock Towards automated icd coding using deep learning.
\newblock \emph{arXiv preprint arXiv:1711.04075}, 2017.

\bibitem[Shickel et~al.(2018)Shickel, Tighe, Bihorac, and
  Rashidi]{shickel2018deep}
Benjamin Shickel, Patrick~James Tighe, Azra Bihorac, and Parisa Rashidi.
\newblock Deep ehr: A survey of recent advances in deep learning techniques for
  electronic health record (ehr) analysis.
\newblock \emph{IEEE journal of biomedical and health informatics}, 22\penalty0
  (5):\penalty0 1589--1604, 2018.

\bibitem[Shore and Johnson(1980)]{shore1980axiomatic}
John Shore and Rodney Johnson.
\newblock Axiomatic derivation of the principle of maximum entropy and the
  principle of minimum cross-entropy.
\newblock \emph{IEEE Transactions on information theory}, 26\penalty0
  (1):\penalty0 26--37, 1980.

\bibitem[Stanfill et~al.(2010)Stanfill, Williams, Fenton, Jenders, and
  Hersh]{stanfill2010systematic}
Mary~H Stanfill, Margaret Williams, Susan~H Fenton, Robert~A Jenders, and
  William~R Hersh.
\newblock A systematic literature review of automated clinical coding and
  classification systems.
\newblock \emph{Journal of the American Medical Informatics Association},
  17\penalty0 (6):\penalty0 646--651, 2010.

\bibitem[Stanford(2002)]{Stanford2002GreedyFA}
Juan F.~Escobar Stanford.
\newblock Greedy function approximation : A gradient boosting machine.
\newblock 2002.

\bibitem[Szegedy et~al.(2015)Szegedy, Vanhoucke, Ioffe, Shlens, and
  Wojna]{DBLP:journals/corr/SzegedyVISW15}
Christian Szegedy, Vincent Vanhoucke, Sergey Ioffe, Jonathon Shlens, and
  Zbigniew Wojna.
\newblock Rethinking the inception architecture for computer vision.
\newblock \emph{CoRR}, abs/1512.00567, 2015.

\bibitem[Tsoumakas and Katakis(2007)]{tsoumakas2007multi}
Grigorios Tsoumakas and Ioannis Katakis.
\newblock Multi-label classification: An overview.
\newblock \emph{International Journal of Data Warehousing and Mining (IJDWM)},
  3\penalty0 (3):\penalty0 1--13, 2007.

\bibitem[Xie and Xing(2018)]{xie2018neural}
Pengtao Xie and Eric Xing.
\newblock A neural architecture for automated icd coding.
\newblock In \emph{Proceedings of the 56th Annual Meeting of the Association
  for Computational Linguistics (Volume 1: Long Papers)}, volume~1, pages
  1066--1076, 2018.

\bibitem[Zhang et~al.(2015)Zhang, Zhao, and LeCun]{zhang2015character}
Xiang Zhang, Junbo Zhao, and Yann LeCun.
\newblock Character-level convolutional networks for text classification.
\newblock In \emph{Advances in neural information processing systems}, pages
  649--657, 2015.

\end{thebibliography}

\newpage
\appendix
\section*{Appendix A.}

\begin{table}[ht!]
\begin{center}
\caption{32 ICD-10 codes and associated descriptions}
\label{tab:32 icd codes}
\begin{tabular}{ll}
\toprule
\begin{tabular}[c]{@{}l@{}}ICD-10 Code\\ \end{tabular} & \begin{tabular}[c]{@{}l@{}}Description\end{tabular} \\ \hline
\begin{tabular}[c]{@{}l@{}}I10\end{tabular}   &Essential (primary) hypertension \\ \hline
\begin{tabular}[c]{@{}l@{}}I50.9\end{tabular}   &Heart failure, unspecified \\ \hline
\begin{tabular}[c]{@{}l@{}}I48.91\end{tabular}   &Unspecified atrial fibrillation \\ \hline
\begin{tabular}[c]{@{}l@{}}I25.10\end{tabular}   &\begin{tabular}[x]{@{}l@{}}Atherosclerotic heart disease of native coronary artery\\without angina pectoris\end{tabular} \\ \hline
\begin{tabular}[c]{@{}l@{}}N17.9\end{tabular}   &Acute kidney failure, unspecified \\ \hline
\begin{tabular}[c]{@{}l@{}}E11.9\end{tabular}   &Type 2 diabetes mellitus without complications \\ \hline
\begin{tabular}[c]{@{}l@{}}E78.5\end{tabular}   &Hyperlipidemia, unspecified \\ \hline
\begin{tabular}[c]{@{}l@{}}N39.0\end{tabular}   &Urinary tract infection, site not specified \\ \hline
\begin{tabular}[c]{@{}l@{}}E78.0\end{tabular}   &Pure hypercholesterolemia, unspecified \\ \hline
\begin{tabular}[c]{@{}l@{}}D64.9\end{tabular}   &Anemia, unspecified \\ \hline
\begin{tabular}[c]{@{}l@{}}E03.9\end{tabular}   &Hypothyroidism, unspecified \\ \hline
\begin{tabular}[c]{@{}l@{}}J18.9\end{tabular}   &Pneumonia, unspecified organism \\ \hline
\begin{tabular}[c]{@{}l@{}}D62\end{tabular}   &Acute posthemorrhagic anemia \\ \hline
\begin{tabular}[c]{@{}l@{}}R65.20\end{tabular}   &Severe sepsis without septic shock \\ \hline
\begin{tabular}[c]{@{}l@{}}F32.9\end{tabular}   &Major depressive disorder, single episode, unspecified \\ \hline
\begin{tabular}[c]{@{}l@{}}F17.200\end{tabular}   &Nicotine dependence, unspecified, uncomplicated \\ \hline
\begin{tabular}[c]{@{}l@{}}D69.6\end{tabular}   &Thrombocytopenia, unspecified \\ \hline
\begin{tabular}[c]{@{}l@{}}Z95.1\end{tabular}   &Presence of aortocoronary bypass graft \\ \hline
\begin{tabular}[c]{@{}l@{}}Z87.891\end{tabular}   & Personal history of nicotine dependence\\ \hline
\begin{tabular}[c]{@{}l@{}}I12.0\end{tabular}   &\begin{tabular}[x]{@{}c@{}}Hypertensive chronic kidney disease with stage 5\\chronic kidney disease or end stage renal disease\end{tabular} \\ \hline
\begin{tabular}[c]{@{}l@{}}R65.21\end{tabular}   &Severe sepsis with septic shock \\ \hline
\begin{tabular}[c]{@{}l@{}}Z79.4\end{tabular}   &Long term (current) use of insulin \\ \hline
\begin{tabular}[c]{@{}l@{}}G47.33\end{tabular}   &Obstructive sleep apnea (adult) (pediatric) \\ \hline
\begin{tabular}[c]{@{}l@{}}J45.909\end{tabular}   & Unspecified asthma, uncomplicated \\ \hline
\begin{tabular}[c]{@{}l@{}}M81.0\end{tabular}   &\begin{tabular}[x]{@{}l@{}}Age-related osteoporosis without current\\ pathological fracture\end{tabular} \\ \hline
\begin{tabular}[c]{@{}l@{}}R56.9\end{tabular}   &Unspecified convulsions \\ \hline
\begin{tabular}[c]{@{}l@{}}N18.6\end{tabular}   &End stage renal disease \\ \hline
\begin{tabular}[c]{@{}l@{}}E66.9\end{tabular}   &Obesity, unspecified \\ \hline
\begin{tabular}[c]{@{}l@{}}R78.81\end{tabular}   &Bacteremia \\ \hline
\begin{tabular}[c]{@{}l@{}}F05\end{tabular}   &Delirium due to known physiological condition \\ \hline
\begin{tabular}[c]{@{}l@{}}E46\end{tabular}   &Unspecified protein-calorie malnutrition \\ \hline
\begin{tabular}[c]{@{}l@{}}E66.01\end{tabular}   & Morbid (severe) obesity due to excess calories\\ \bottomrule
\end{tabular}
\end{center}
\end{table}

\begin{table}[]
\caption{Samples of Prediction Interpretability from different modalities.  Samples are displayed as in descending order and their rankings are shown in brackets. Due to space limit, only top 3 unstructured text results are included, except those mentioned in \ref{interp_discussion}.}
\label{interpretability demonstration}
\begin{tabular}{@{}cl@{}}
\toprule
\multicolumn{2}{c}{\textbf{Interpretability Demonstration}} \\ \midrule
\multicolumn{2}{l}{D64.9 Anemia, Unspecified} \\ \midrule
Unstructured Text & \begin{tabular}[c]{@{}l@{}}`Right Ventricular' (1)\\ `Postoperative Pleural Effusions, Anemia, Acute Renal (2)\\ 'nosocomial pneumonia. Cultures eventually grew out mrsa' (3)
\end{tabular} \\
LABEVENTS & `Hematocrit'(1), `Hyaline Casts'(2), `pO2'(3), `Hemoglobin'(10) \\
PRESCRIPTION & \small{`Vancomycin'(1), `Heparin'(2), `Sodium Chloride 0.9\% Flush'(3)} \\ \midrule
\multicolumn{2}{l}{E11.9 Type 2 diabetes mellitus without complications} \\ \midrule
Unstructured Text & \begin{tabular}[c]{@{}l@{}}`Medical History: Copd Obesity DM II' (1) \\ `nebs changed to inhalers. Pts DM' (2)\\ `COPD, DM, Tobacco Abuse, Obesity, Iron Deficiency Anemia' (3)\\
`Metformin 850 mg po' (5)
\end{tabular} \\
LABEVENTS & `Glucose'(1), `Basophils'(2), `Lactate Dehydrogenase (LD)'(3) \\
PRESCRIPTION & \small{`Insulin'(1), `Metformin'(2), `Glyburide'(3),'Humulin-R Insulin'(6)} \\ \midrule
\multicolumn{2}{l}{I10 Essential (primary) hypertension} \\ \midrule
Unstructured Text & \begin{tabular}[c]{@{}l@{}}`coumadin, HTN, COPD, Hepatocellular carcinoma' (1)\\ `diagnostic thoracentesis, urinalysis negative, urine cx' (2)\\ `subsequent enucleation, stent in pancreas' (3)
\end{tabular} \\
LABEVENTS & `Creatinine'(1), `Monocytes'(2), `Myelocytes'(3) \\
PRESCRIPTION & \small{`Atenolol'(1), 
`Sodium CHloride 0.9\% Flush'(2), `Albuterol-Ipratropium'(3)}\\ 
\midrule
\multicolumn{2}{l}{I50.9 Heart failure, unspecified} \\ \midrule
Unstructured Text & \begin{tabular}[c]{@{}l@{}}`Lisinopril' (1), `Anxiety/Depression, CKD, HLD, Obesity, HTN' (2)\\ `Chronic Systolic Heart Failure' (3)
\end{tabular} \\
LABEVENTS & `Troponin T'(1), `Urea Nitrogen'(2), `Heart\_rate'(3) \\
PRESCRIPTION & \small{`Furosemide'(1), `Carvedilol'(2), `Topiramate (Topamax)'(3) ,`Lisinopril'(4)}
\\ \midrule
\multicolumn{2}{l}{N17.9 Acute kidney failure, unspecified} \\ \midrule
Unstructured Text & \begin{tabular}[c]{@{}l@{}}`Acute Kidney Injury: Multifactorial etiology' (1), `Renal failure' (2)\\ `Hands, wrists, elbows, shoulders, forearm developed a' (3)
\\ `IVDU, states he used heroin, barbiturates, cocaine' (5)\\ 


\end{tabular} \\
LABEVENTS & `Creatinine'(1), `Urea Nitrogen'(2), `Monocytes'(3) \\
PRESCRIPTION & \small{`NaHCO3'(1), `Iso-Osmotic Dextrose'(2), `Phytonadione'(3)} \\
\midrule
\multicolumn{2}{l}{}
\end{tabular}
\end{table}

\end{document}